# Autoencoding Features for Aviation Machine Learning Problems

Liya Wang,[1] Panta Lucic,[2] Keith Campbell,[3] and Craig Wanke[4]

*The MITRE Corporation, McLean, VA, 22102, United States*

The current practice of manually processing features for high-dimensional and heterogeneous aviation data is labor-intensive, does not scale well to new problems, and is prone to information loss, affecting the effectiveness and maintainability of machine learning (ML) procedures. This research explored an unsupervised learning method, autoencoder, to extract effective features for aviation machine learning problems. The study explored variants of autoencoders with the aim of forcing the learned representations of the input to assume useful properties. A flight track anomaly detection autoencoder was developed to demonstrate the versatility of the technique. The research results show that the autoencoder can not only automatically extract effective features for the flight track data, but also efficiently deep clean data, thereby reducing the workload of data scientists. Moreover, the research leveraged transfer learning to efficiently train models for multiple airports. Transfer learning can reduce model training times from days to hours, as well as improving model performance. The developed applications and techniques are shared with the whole aviation community to improve effectiveness of ongoing and future machine learning studies.

## I. Nomenclature

| | | |
|---|---|---|
| $X$ | = | $[x_1, x_2, x_3, ..., x_n]$, input data |
| $X'$ | = | reconstructed output of input $X$ |
| $Z$ | = | bottleneck, compressed low dimensional representation of input $X$ |
| $g_\phi$ | = | encoding function parameterized by $\phi$ |
| $f_\theta$ | = | decoding function parameterized by $\theta$ |
| $MAE$ | = | mean absolute error, the metric to measure the reconstruction error between $X$ and $X'$ |
| $\delta$ | = | anomaly detection threshold of mean absolute error (MAE) |

## II. Introduction

With recent advancements in computing power, big data, and algorithms, machine learning (ML) has gained increasing prominence in many areas such as image recognition, facial recognition, natural language processing, self-driving cars, etc. Compared to traditional rule-based methods where the rules are defined by experts, ML algorithms can automatically learn rules from complex data. The learning ability, adaptability, and flexibility of ML methods can bring multiple benefits such as reducing subjective errors and labor cost.

Aviation researchers have begun to adopt ML to advance the area. For example, Smith et al. [1] did a survey of different technology forecasting techniques for complex systems and identified machine learning useful for providing estimates for future technology predictions; Wang et al. [2] applied logistic regression for an unstable approach risk prediction at a specific distance location to the runway threshold. In addition, a variety of ML algorithms were used to predict aviation demand ( [3]); Kim et al. [4] proposed long-short term memory (LSTM) to predict flight delays.

---

[1] Data Scientist, Lead, P223— Operation Performance
[2] Computer Science, Lead, L252— Transportation Performance & Economic Analytics
[3] System Engineering, Principal, L254— Transportation Data Analytics
[4] Division Chief Engineer, CAASD

Even with the aforementioned ML applications, the aviation domain, compared to other domains, has still been relatively slow to adopt ML research. Although we have abundant aviation data such as flight tracks, weather, safety reports, etc., the data are seldom in a ready-to-use format for the application of machine learning algorithms. In addition, those data are high-dimensional and heterogeneous. For example, a flight track can have thousands of time sequence points, and different tracks may have a different number of points. The current practice of manually processing features is labor-intensive, does not scale well to new problems, and is prone to information loss, affecting the effectiveness and maintainability of ML procedures. For that, this study proposed an unsupervised learning method, autoencoder, to automatically extract effective features. The research also explored variants of autoencoders, in an effort to force the learned representations of the input to assume useful properties. A flight track anomaly detection autoencoder has been developed to demonstrate the versatility of the technique.

The remainder of the paper is organized as follows: Section III gives a short introduction about autoencoders, and Section IV presents the flight track anomaly detection autoencoder. Section V describes data collection, with results will be shown in Section VI. Section VII describes transfer learning, used to expeditiously train models for multiple airports. Finally, we conclude in Section VIII.

### III. Autoencoder

An autoencoder is a neural network designed for the task of representation learning in an unsupervised way. It tries to reconstruct the original input while compressing the data in the process so as to discover a more efficient and compact representation (Fig. 1) . The idea was originated in the 1980s, and later promoted by the seminal paper of Hinton and Salakhutdinov ( [5]).

It consists of two parts:

- **Encoder**: Compresses the original high-dimension input into the latent low-dimensional representation.

- **Decoder**: Attempts to reconstruct the data from the latent representation.

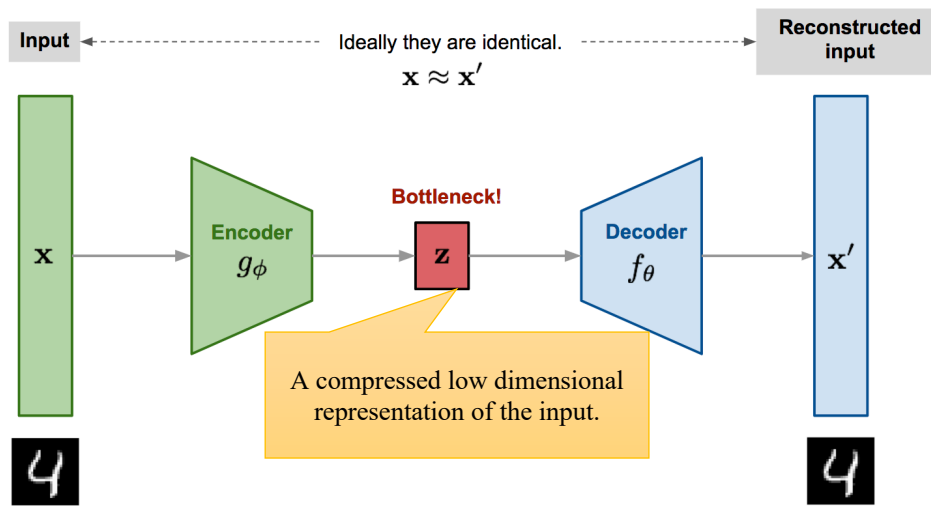

Fig. 1. Illustration of autoencoder model architecture

Fig. 1 [6] illustrates the process and architecture of the autoencoder model. By giving the input $X$ to the encoder, and with the decoder outputting $X'$, the neural network is trained to make $X$ and $X'$ nearly identical, with the difference measured by the reconstruction error. The bottleneck $Z$ is a compressed low dimensional representation of input $X$, which can then be used to support ML modeling. In our study, mean absolute error (MAE) is used to measure the reconstruction error (Eq. 1).

$$\text{MAE}= \sum_{i=1}^{n}(x_i - x'_i)/n \tag{1}$$

## IV. Flight Track Anomaly Detection Autoencoder

Flight-track data are commonly used in the aviation domain for studies such as safety risk prediction, traffic flow management, and procedure-based navigation route design. However, the data have two features making them difficult to use in machine learning algorithms: high-dimensionality (numerous time series data points) and heterogeneity. For example, in MITRE threaded track data [7], a data point is recorded every 1 to 4 seconds in the terminal area for each flight. In addition, each flight travels different routes. Hence, each flight track is a variable-length time series. It is essential to reduce dimension before feeding long time-sequence data into a machine learning algorithm to train models. For this reason, the autoencoder is an excellent option.

Various types of autoencoders, including anomaly detection autoencoders, have been designed to tackle different types of problems. Detecting anomalies is key for almost any quantitative analysis. In the process of producing flight data, anomalies may be introduced inadvertently from many sources and hidden in many dimensions. It is difficult for analysts to enumerate all the data cleaning rules, which typically depend on the analysts' experience and expertise with the data. All the above reasons provide motivation for creating an artificial intelligent (AI) agent to support more efficient and automated data deep cleaning, thereby reducing analyst workload.

To build the flight track anomaly detection autoencoder, we have designed a three-step procedure:

**Step 1.** Feed preliminary "normal" track data to train autoencoder models.

The more details about preliminary "normal" will be explained in Section V. This step will make the autoencoder remember the normal or expected pattern.

**Step 2.** Setup reconstruction error threshold to automatically detect anomalies.
When unusual data is present, the autoencoder will have trouble reproducing it and the subsequent reconstruction error will be large. When the error is above some threshold, an alarm will be raised.

**Step 3.** Test and validate.
If necessary, repeat 2 and 3 until the required performance is achieved.

While building our neural network model, an important research question is which type of neurons we should choose. Our track data are time series data. For that, two options are considered: convolution neural network (CNN) or recurrent neural network (RNN). 1D CNN (e.g., CONV1D in tensorflow [8]) is designed for sequence data [9]. Given that our track data have long time sequences (e.g., over 2000 points for some tracks), and RNN has a gradient vanishing problem when dealing with long sequences, CONV1D is chosen for our encoder to compress the data, and CONV1DTranspose to decompress the data (Fig. 2) ( [10]).

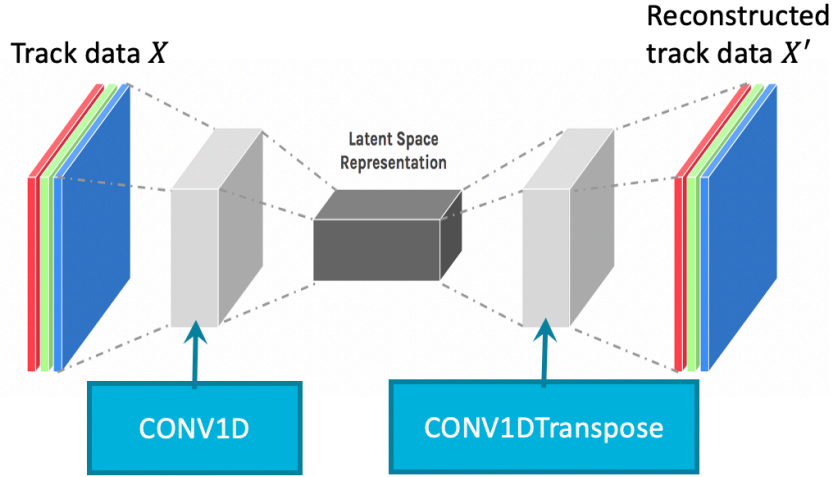

**Fig. 2. Flight track anomaly detection autoencoder**

Fig. 3 shows the designed training model architecture used in this effort, which depicts all the necessary data preprocessing, train-test data splitting, and modeling work carried out. Given the expensive joining process and large memory consumption of flight track data, the data was iteratively processed month by month. Additionally, we used pyspark for this task in order to take advantage of parallel processing. We then followed our designed procedures to obtain the train and test datasets. After that, three-step modeling procedure was carried out to train the model.

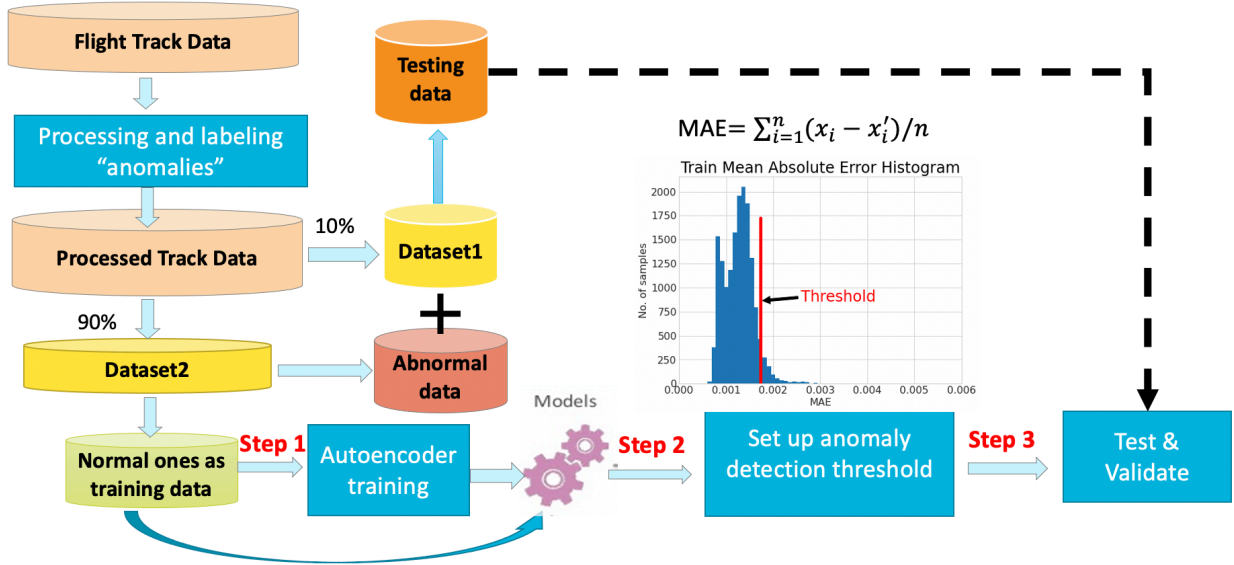

**Fig. 3. Flight track anomaly detection training architecture**

## V. Data Collection

For our research, we identified five datasets in MITRE's Transportation Data Platform (TDP) [7], where a suite of surveillance data and supporting datasets including infrastructural and operational information are provided. Table 1 lists the five datasets ad summarizes their functionalities. Fig. 4 shows the extract, transform, load (ETL) process used in this effort.

**Table 1. Transportation Data Platform (TDP) Data Sources**

| Dataset | Function |
|---|---|
| **National Flight Data Center (NFDC)** | Airport and runways infrastructure data table with information such as:<br>• Airport latitude, longitude, elevation<br>• Runway threshold latitude, longitude, and elevation |
| **Threaded Track** | Surveillance data table with information such as:<br>• Time, latitude, longitude, altitude, course, speed etc.<br>• Aircraft type |
| **Airport Runway Assignment** | Identifies probable landing runway based on track data points |
| **Missed Approach** | Identifies flights with missed approach |
| **Aircraft Type** | Aircraft type information such as helicopters, military ones, and unmanned aircraft system (UAS) derived from FAA's Order 7360 [11] |

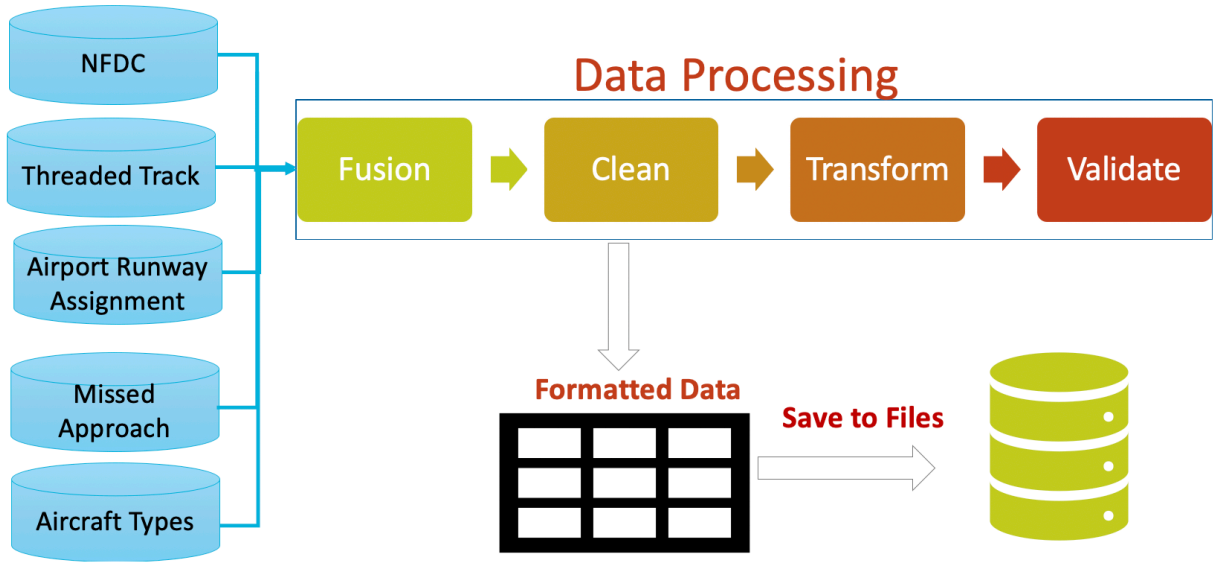

**Fig. 4. Data collection process**

For the training process, we processed arrival track segments in terminal airspace, or 40 nautical miles (NM) from the runway threshold. Fig. 5 provides examples of our processed sample track data at Las Vegas McCarran International Airport (LAS). The nominal arrival patterns include a gradual reduction of both altitude and speed, as depicted in Fig. 6.

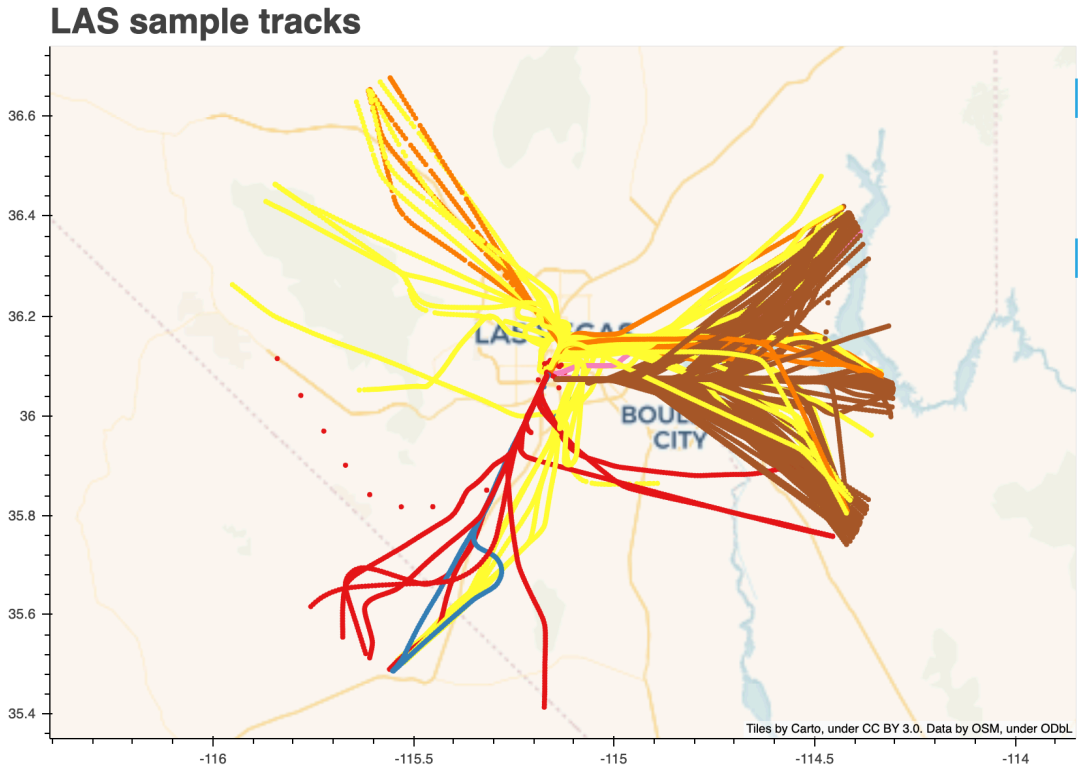

**Fig. 5. Processed sample track data for LAS airport**

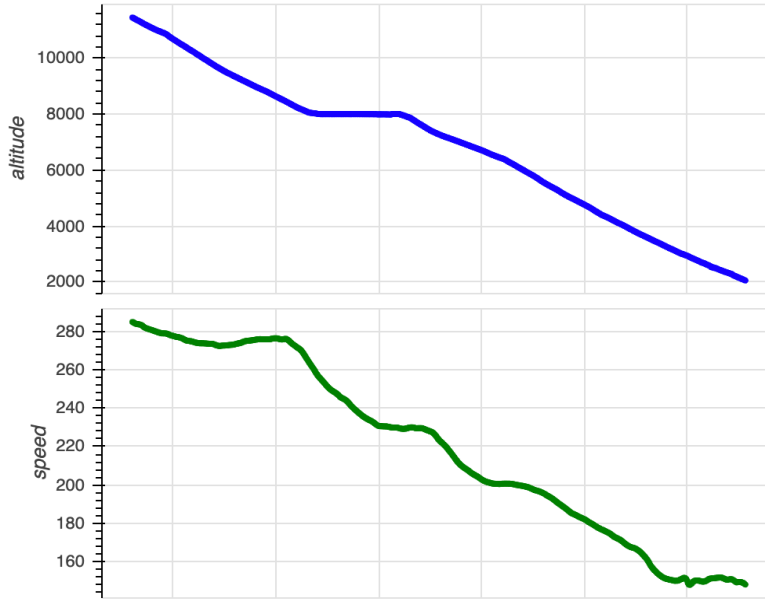

**Fig. 6. Normal pattern of an arrival track**

The TDP dataset does not have a label to identify whether a track is "normal" or not. To identify preliminary "normal" track segments, we leveraged the TDP-provided track characteristics and experts' opinions on what is considered "normal". The rules reflect a focus on commercial and general aviation flights, so other types of flights with different characteristics, such as helicopter traffic, are excluded. If a track has any of the characteristics as shown in Fig. 7, we label them as potential anomalies and filter them out of the dataset. Next, the remaining preliminary "normal" data are used to train the autoencoder. The filtering step can help the autoencoder quickly learn the normal arrival patterns with less noisy data. After that, we then selected the speed and altitude of each track as features to train our autoencoder.

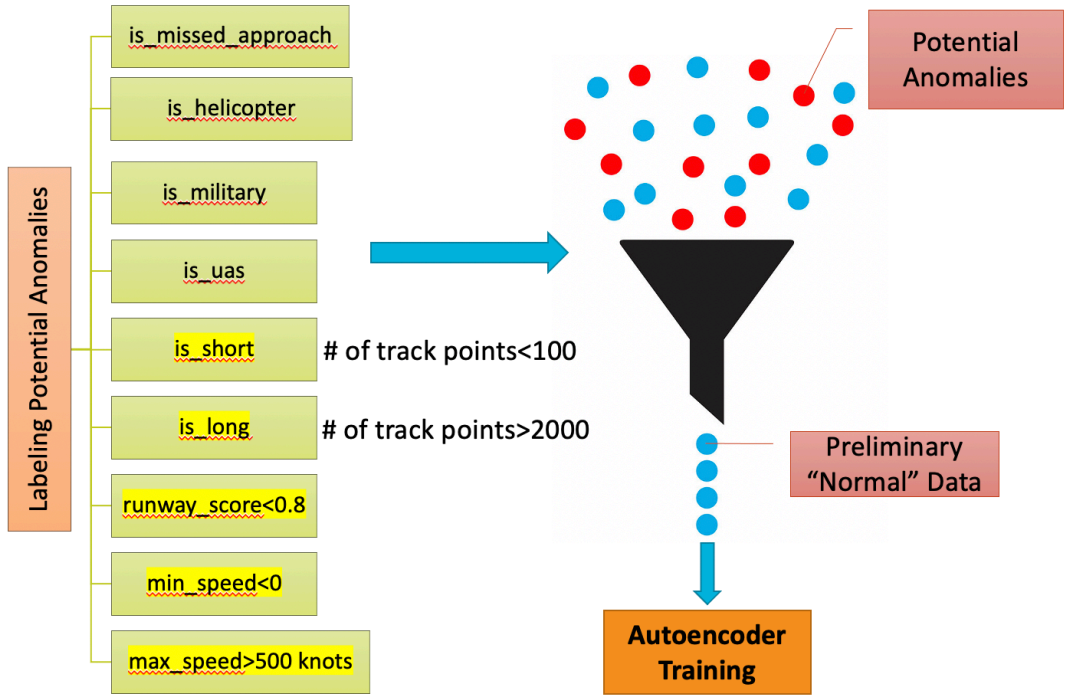

**Fig. 7. Labeling anomalies for the track data**

## VI. Anomaly Detection Results

In this section, we present some of our feature extraction and anomaly detection results.

### A. Feature Extraction Results

Fig. 8 presents a randomly selected normal track, where Fig. 8 (a) shows the map of the track while Fig. 8 (b) compares altitude and speed of original track versus the reconstructed tracks. The red lines represent data from the original track, and the blue lines represent the reconstructed tracks. Where the points closely coincide, the lines become purple. For the example presented in Fig. 8, the two lines agree quite well with each other. This example demonstrates how the autoencoder can successfully learn effective latent representation for a normal track.

In contrast, Fig. 9 shows an abnormal track. Since the autoencoder cannot determine the abnormal behavior of this track (see Fig. 9 (b)), the red lines and blue lines do not align. In this case, it appears that the observed anomaly is caused by noisy data received from the FAA's Traffic Flow Management System (TFMS).

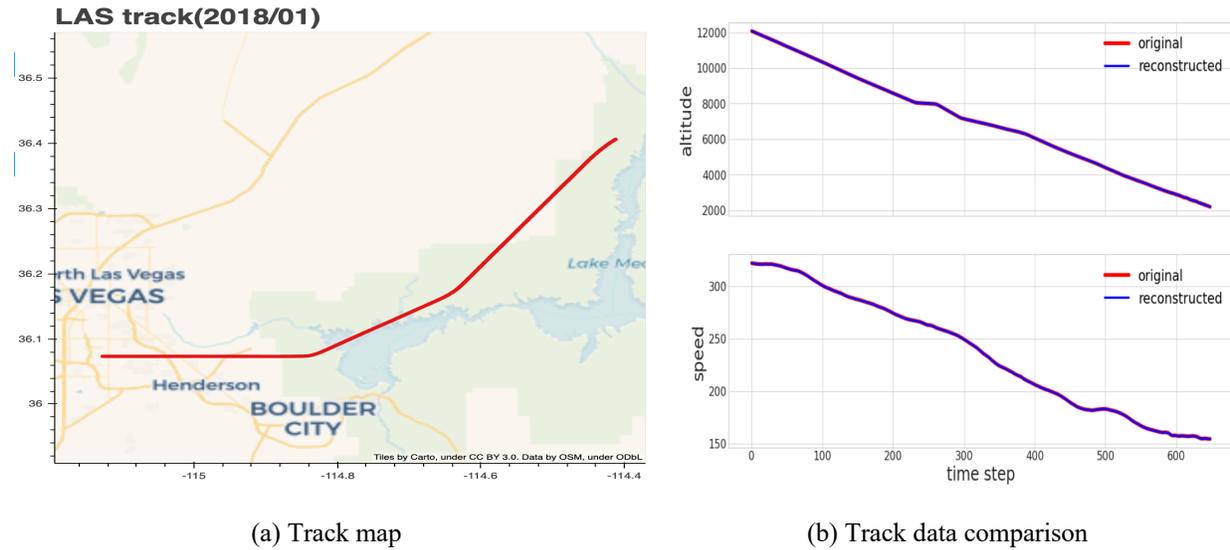

(a) Track map  (b) Track data comparison

**Fig. 8. Sample normal track feature extraction**

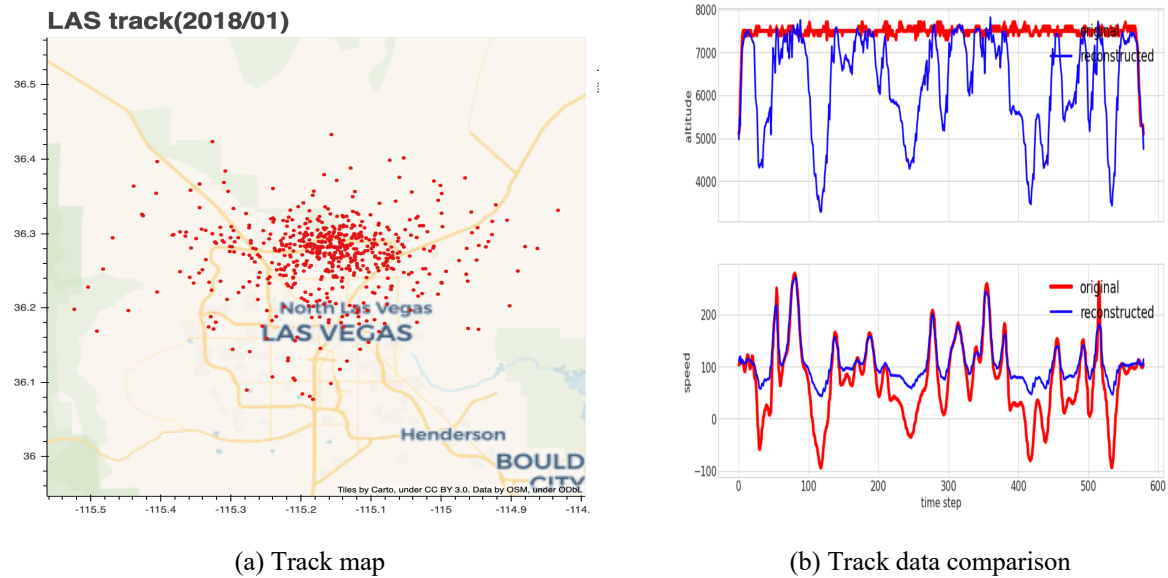

(a) Track map  (b) Track data comparison

**Fig. 9. Sample abnormal track feature extraction**

We also examined the tracks whose MAEs are near the threshold boundary to ensure that normal tracks are resembling (see Fig. 8 (b)) and that abnormal tracks are not alike (see Fig. 9 (b)). Otherwise, fine tuning of our neural network model (or adjusted MAE thresholds, targeting expected patterns) is required.

B. Anomaly Detection Results Summary

Fig. 10 summarizes LAS anomaly statistics for January 2018 traffic. On average, LAS has approximately 20,000 arrival operations per month. Fig. 10 (a) shows that 9.5% of LAS tracks are anomalous, with 42% of these anomalies associated with helicopters (see Fig. 10 (b)). In addition, a positive correlation was observed between the percentage of helicopters and the percentage of anomalies. Furthermore, our AI agent can complete 5 years (2015-2019) of track data anomaly detection for LAS in less than 5 hours, using nominal CPU processing. Collectively, these observations fully demonstrate that our AI agent can efficiently deep clean data and do so in an automated manner that significantly reduce analyst workload.

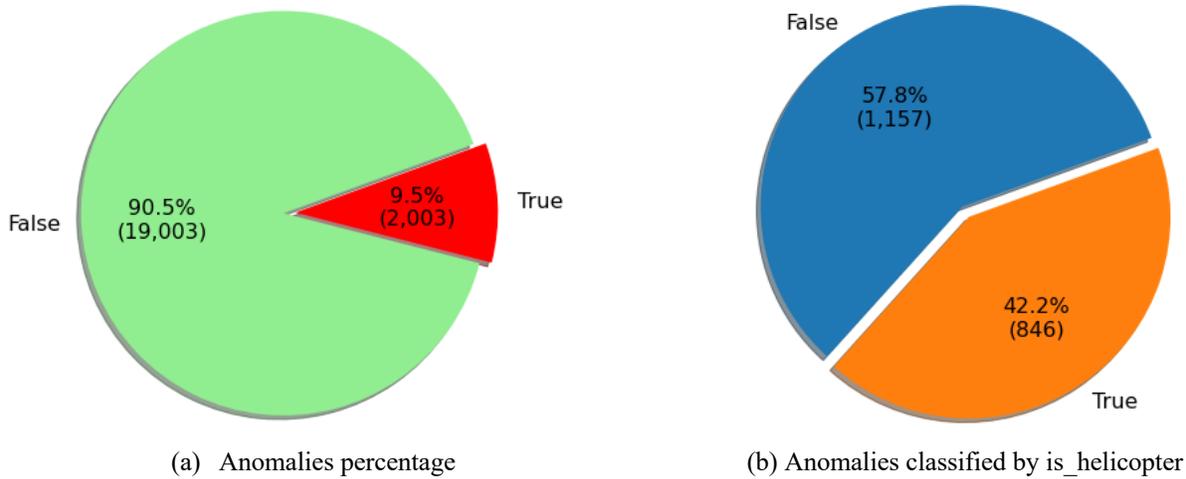

(a) Anomalies percentage          (b) Anomalies classified by is_helicopter

**Fig. 10. LAS flight track anomaly results for January 2018**

C. Anomaly Classification

For better understanding about the detected anomalies, identified anomalous tracks were grouped into categories. Fig. 11 depicts our classification method. First, we classify anomalies by aircraft weight class. In each class, we classify them into non-notable versus notable anomalies. The non-notable anomalous tracks are related to known patterns such as arrivals from internal airports within terminal (40 NM) airspace, or medical, or television, or touring helicopters. The notable anomalous tracks are those related to data quality issues or risky operations. Six types of notable anomalous tracks were identified, and examples are subsequently provided.

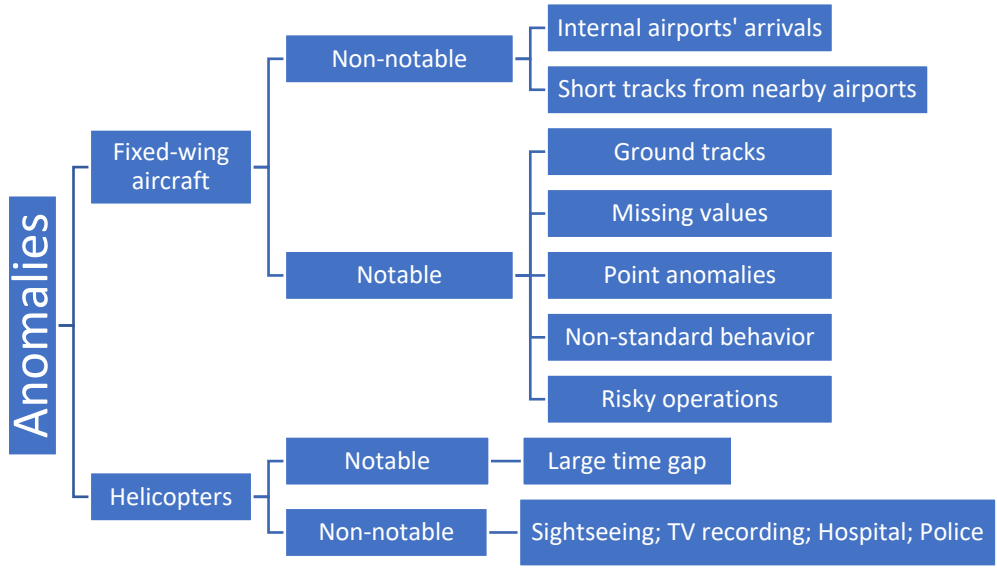

Fig. 11. Anomalies classification

### D. Sample Notable Track Anomalies

This section depicts examples of the six notable anomaly types. Fig. 12 provides an example of a ground track anomaly. This anomaly occurs when different data sources provide insufficient information to accurately assemble full origin to destination trajectories, leaving an unassigned trajectory fragment. Identification of these anomalies and their in-depth analyses are expected to lead to improvement of the algorithms currently used to assemble the trajectories.

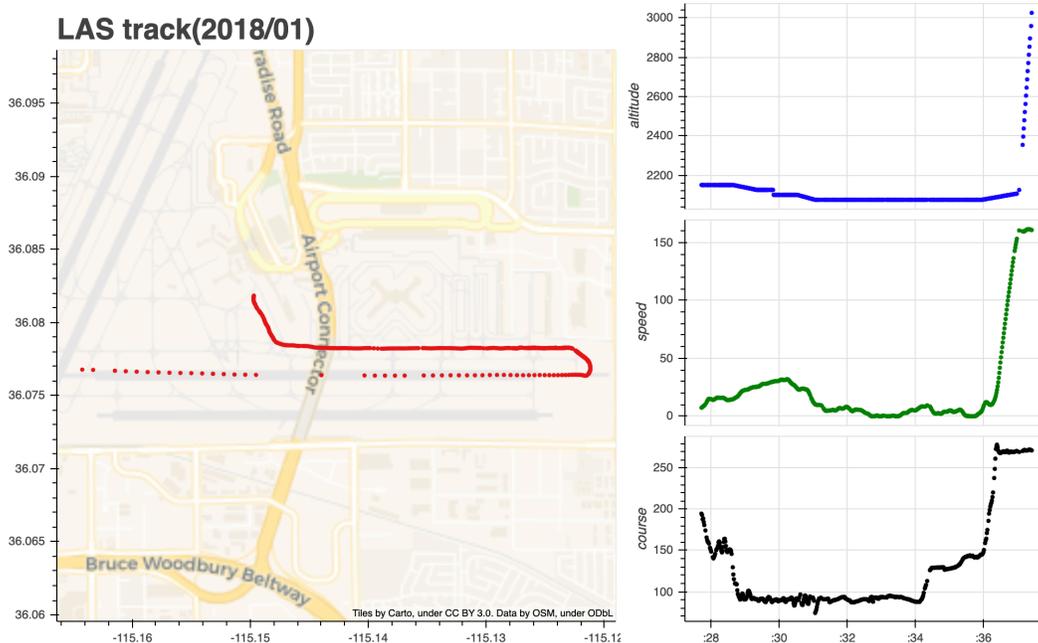

Fig. 12. Ground track anomaly example

Fig. 13 and 14 show examples of track point anomalies. In Fig. 13, the anomaly is caused by noise in altitude data. In Fig. 14, the anomaly is caused by noisy speed data.

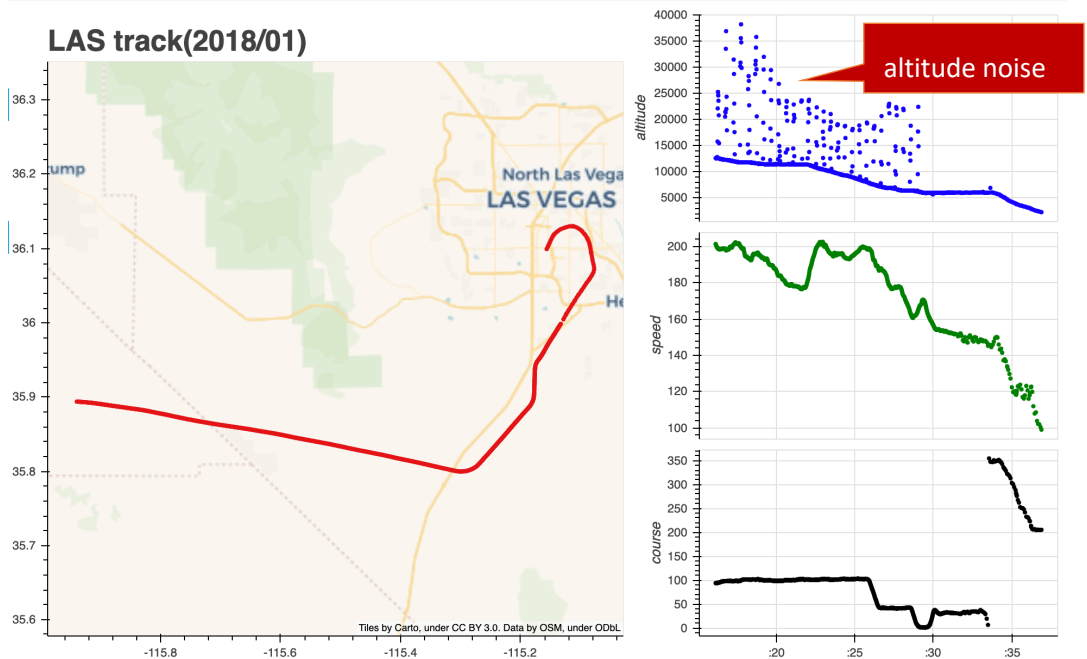

**Fig. 13. Point anomaly example 1**

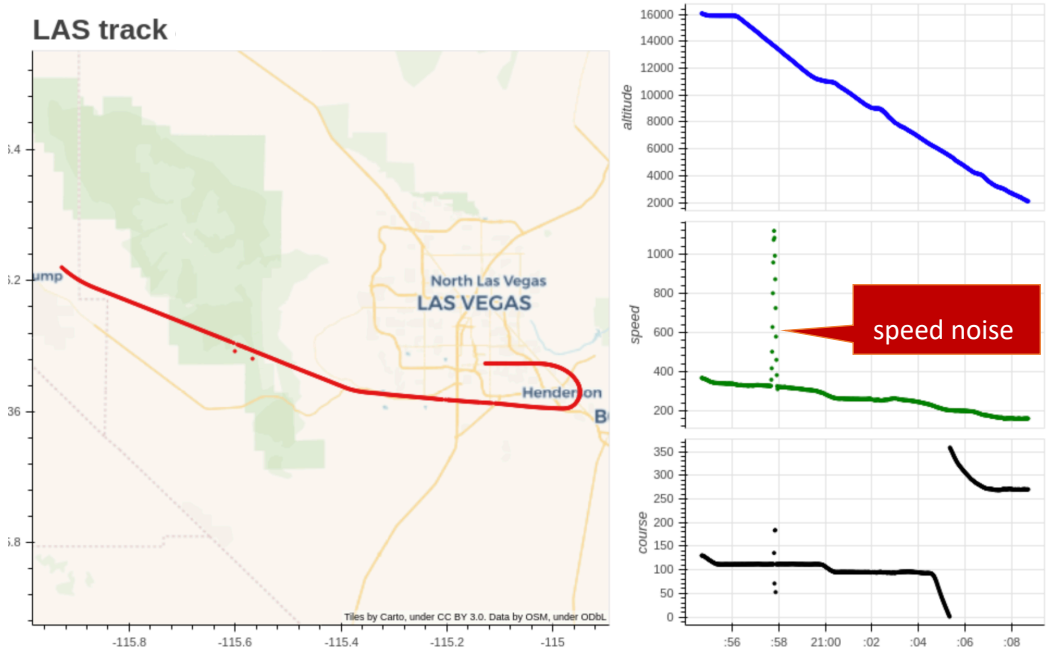

**Fig. 14. Point anomaly example 2**

Fig. 15 shows an anomaly caused by missing data. The track has a large track section with missing altitude information (see trajectory segment in black in Fig. 15).

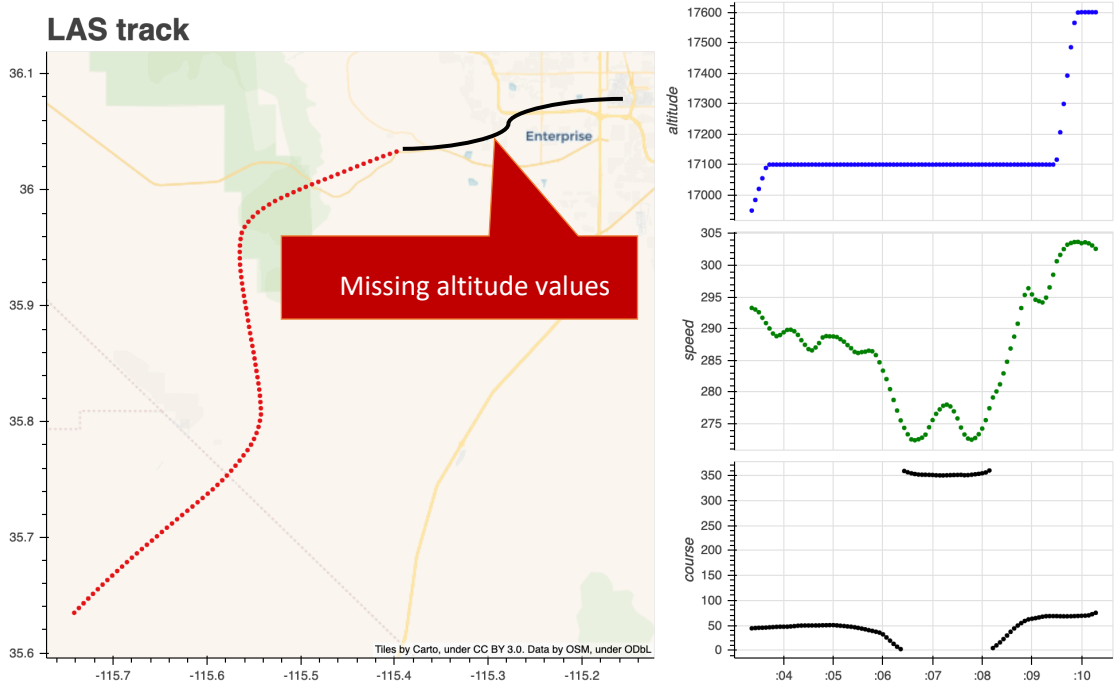

**Fig. 15. Missing altitude for a large piece of track anomaly example**

Fig. 16 shows a non-standard operation anomaly, existing because the track does not include a takeoff roll, and also because the flight lands after just 5 airborne minutes. These behaviors are rarely seen for normal operations. This may represent a track fragment incorrectly separated from the remainder of the flight.

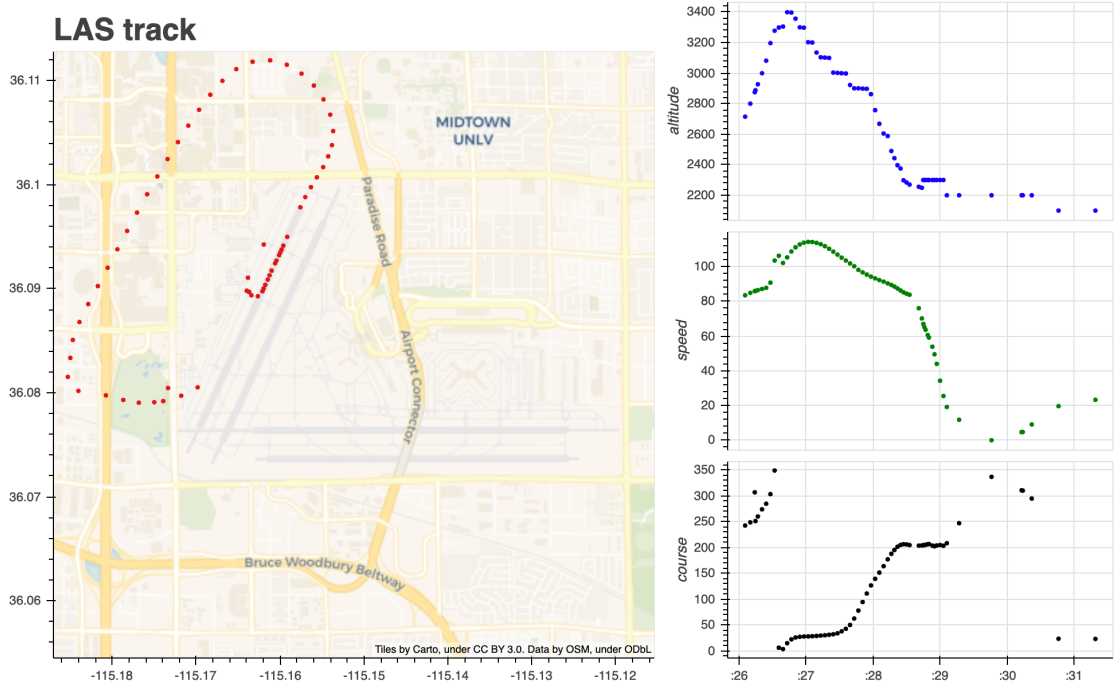

**Fig. 16. Non-standard operation anomaly example**

Fig. 17 shows an anomaly indicative of a risky operation, where the "fixed-wing" flight operated totally in a small fixed-base operator (FBO) and reached 51,000 feet while staying inside the FBO. Compared to normal operations, this is a very steep climb and descent. This flight also does not have takeoff or landing rolls, and overall does not look like a realistic track. With consideration that Van Nuys Airport (VNY) is (a) listed as a Gulfstream service center, (b) supposedly there are at least 130 aircraft present there, and (c) the service ceiling for these aircraft is 51,000 feet, aviation domain experts believe this flight may have been conducting transponder tests within the FBO.

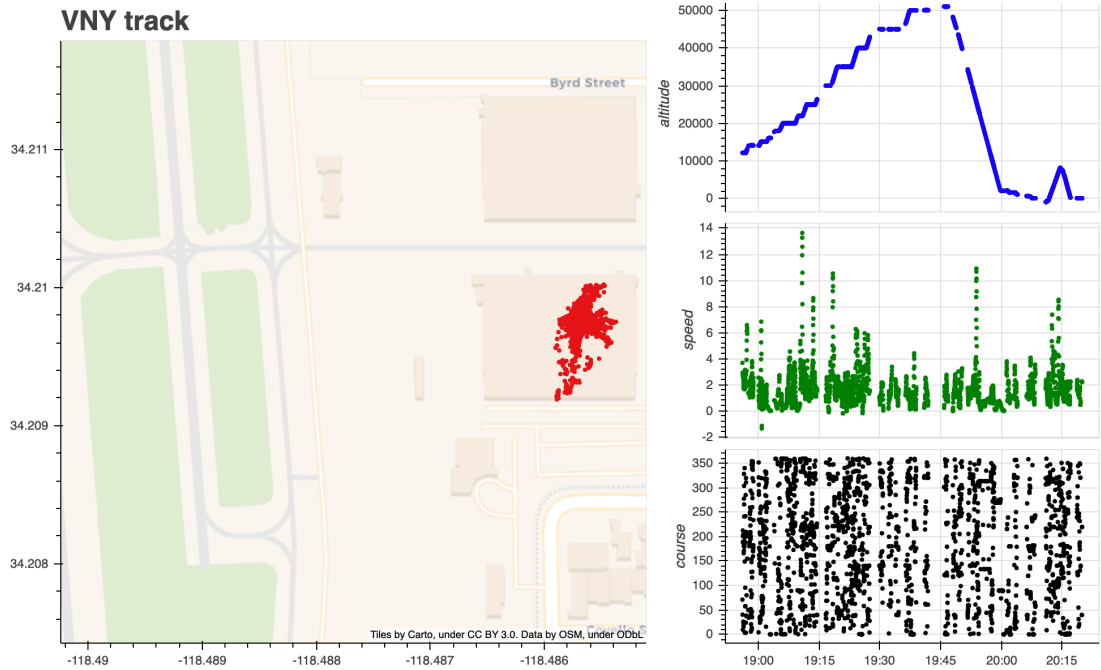

**Fig. 17. Risky operation anomaly example**

Fig. 18 presents a large time gap anomaly for a helicopter track. The gap is over one hour in length, which is much larger than the expected time gap (less than 12 seconds) between two consecutive track points. This may indicate two different flights have been incorrectly combined. Many anomalies of this type indicate the need to re-examine the processing of helicopter data/tracks.

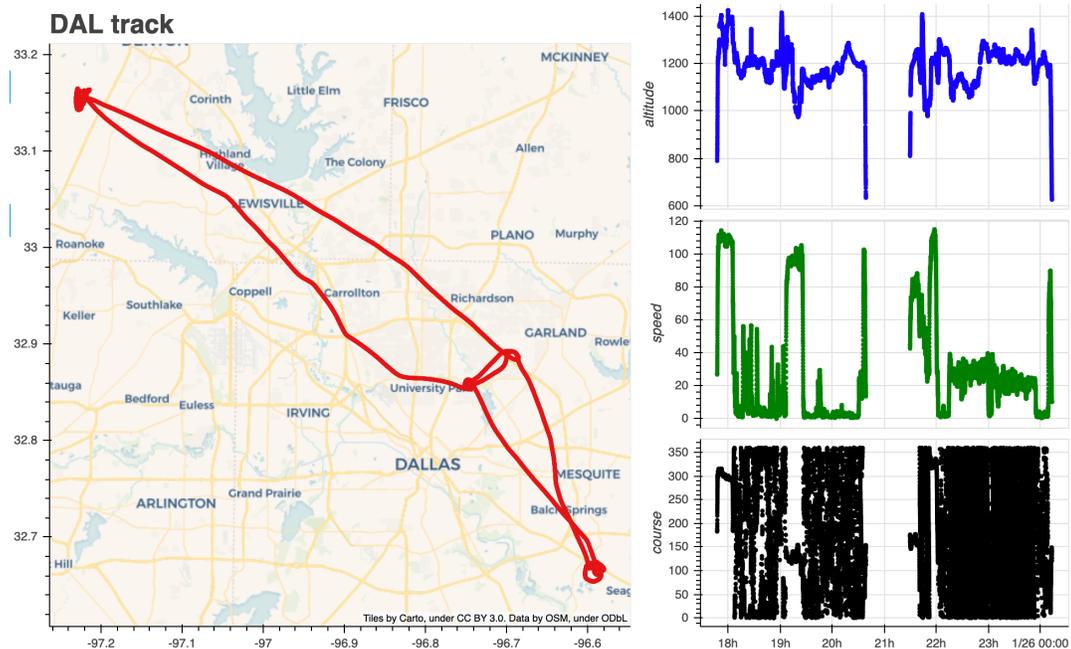

**Fig. 18. Large time gap anomaly example**

Through presenting all the above aviation track anomaly examples, it is important to emphasize that "anomaly" should not incur a negative connotation. On the contrary, many insights about the data considered foundational to our aviation and air traffic analyses are gained by anomaly identification, alerting us on where and how to improve our data ecosystem and processing algorithms.

## VII. Transfer Learning

To account for unique properties of nominal arrival tracks to different airports, each airport needs to have its own ML model. We have leveraged transfer learning to train multiple models. Transfer learning uses knowledge from a previously trained model and applies it to a related task (Fig. 19). For example, knowledge from model A could be applied when trying to train the model B. Transfer learning is considered to be the next driver of commercial ML success after supervised learning [12].

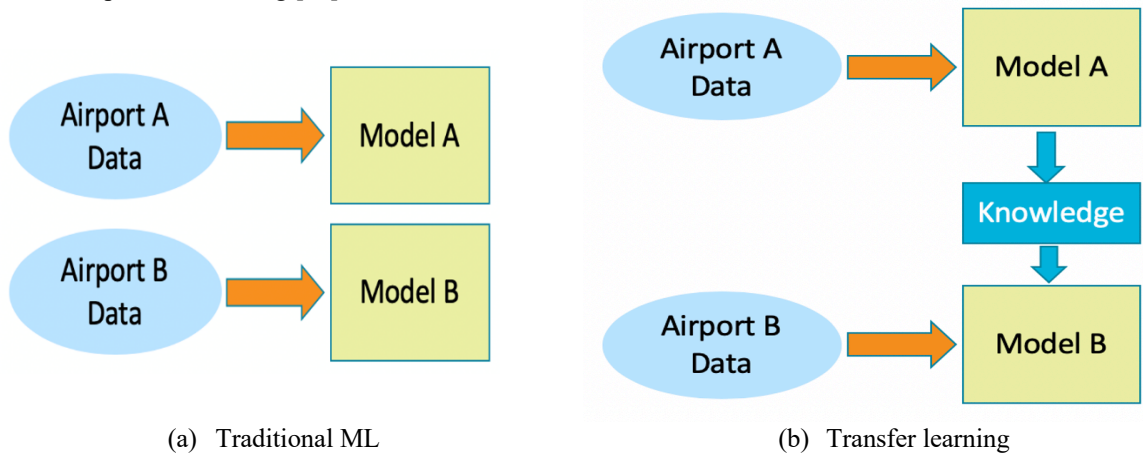

(a) Traditional ML  (b) Transfer learning

**Fig. 19. Traditional ML vs transfer learning**

In transfer learning, there are two common approaches: 1) developed model approach, and 2) pre-trained model approach [13]. The latter approach is commonly used in the field of deep learning. However, in the aviation domain, we do not have matured models like AlexNet [14] in the computer vision field or Bert [15] in the NLP field. Therefore, we employed the first approach, which can be described as follows:

1. **Select Source Task**. Select a predictive problem with an abundance of data.
2. **Develop Source Model**. Next, train model for the first task.
3. **Reuse Model**. The source model can then be used as the starting point for a model on the target task. We froze the first layer of source model and then reused it.
4. **Tune Model**. The other layers of the source model were refined with the target task dataset.

Applying this approach in our work, we first trained a source model for LAS, and then leveraged it to train target models for six other airports (Fig. 20). Table 2 lists the airports for which we have trained models. We found that transfer learning can reduce training times from days to hours, and the improved performance can also be observed through the loss function optimization process.

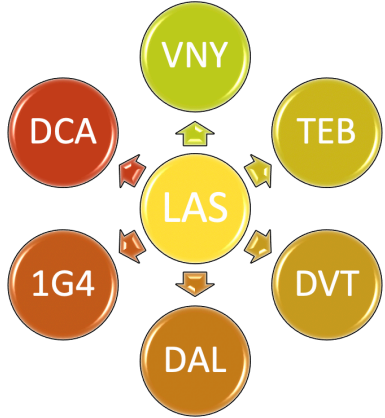

**Fig. 20. Transfer learning for flight track anomaly detection autoencoder**

Table 2. Airports for which trained models were developed

| FAA Code | Airport Name |
|---|---|
| LAS | McCarran International Airport |
| DAL | Dallas Love Field |
| DCA | Ronald Reagan Washington National Airport |
| DVT | Phoenix Deer Valley Airport |
| TEB | Teterboro Airport |
| VNY | Van Nuys Airport |
| 1G4 | Grand Canyon West Airport |

## VIII. Conclusion

This paper described the development of autoencoders to automatically extract features for high-dimensional and heterogeneous aviation data. We also developed a flight track anomaly detection autoencoder for flights arriving at a specific airport to demonstrate the versatility of the technique. The results of this research show that an autoencoder not only can learn effective representation for flight track data, but can also efficiently deep clean data, thereby reducing workload for the analysts. Moreover, the research leveraged transfer learning to efficiently train models for multiple airports. Transfer learning can reduce model training times from days to hours, as well as improve overall model performance.

## Acknowledgments

We thank the following MITRE colleagues: Clark Lunsford, Dr. Alex Tien, Mike Robinson, Dan Larson, Joe Hoffman, Karl Mayer, Van Hare, Eric Zakrzewski, Adric Eckstein, Gaurish Anand, Evan McClain, Paul Diffenderfer, and Dr. Lakshmi Vempati for the valuable discussions and insights.